\pdfoutput=1
\documentclass[11pt]{article}

\usepackage[final]{acl}

\usepackage{times}
\usepackage{latexsym}
\usepackage[T1]{fontenc}
\usepackage[utf8]{inputenc}
\usepackage{microtype}
\usepackage{inconsolata}
\usepackage{graphicx}
\usepackage{amsmath}
\usepackage{amssymb}
\usepackage{booktabs}
\usepackage{multirow}
\usepackage{tabularx}
\usepackage{array}
\usepackage{listings}
\defcitealias{inep2024enemProvas}{INEP 2024}

\newcommand{\methodname}{HACHIMI-II}

\title{\methodname: From Role-Play to Learner Behavior Modeling\\for Faithful Student Simulation}

\author{
 Yilin Jiang\textsuperscript{1,2},
 Fei Tan\textsuperscript{1}\thanks{Corresponding author.},
 Xuanyu Yin\textsuperscript{1},
 Jing Leng\textsuperscript{1},
 Aimin Zhou\textsuperscript{1,3}\\
 \textsuperscript{1}East China Normal University, Shanghai, China\\
 \textsuperscript{2}The Hong Kong University of Science and Technology (Guangzhou), Guangzhou, China\\
 \textsuperscript{3}Shanghai Innovation Institute, Shanghai, China\\
 \texttt{ftan@mail.ecnu.edu.cn}}

\begin{document}
\maketitle

\begin{abstract}
Simulated students let researchers test tutoring systems and teaching strategies before real learners are available, but only if the simulation reproduces where a particular learner succeeds and fails. The common approach asks a large language model to role-play a student, but a role does not model what this learner knows or will do next, so the model's own competence fills the gap. Simulated students perform at a level set by the model, differ little from one another, and seldom make the real learners' errors. We model the learner's behavior and recast student simulation as \textbf{third-person prediction of the learner's next action}. \methodname{} builds a student model from the learner's history, infers the learner's state for the current item or tutor turn, and compiles it into a behavior harness that constrains the next action. On three item-response datasets, the Wasserstein-1 distance between simulated and real student ability is \textbf{2.4--8.4$\times$} smaller than that of the strongest baseline. On two dialogue datasets, \methodname{} reproduces real student errors with the same knowledge type, locus, and mechanism \textbf{1.8--5.1$\times$} more often than the strongest baseline, and ablations trace the gains mainly to the explicit learner state. Predicting the learner makes capable models usable as simulated students.
\end{abstract}

\begin{figure*}[t]
\centering
\includegraphics[width=\textwidth]{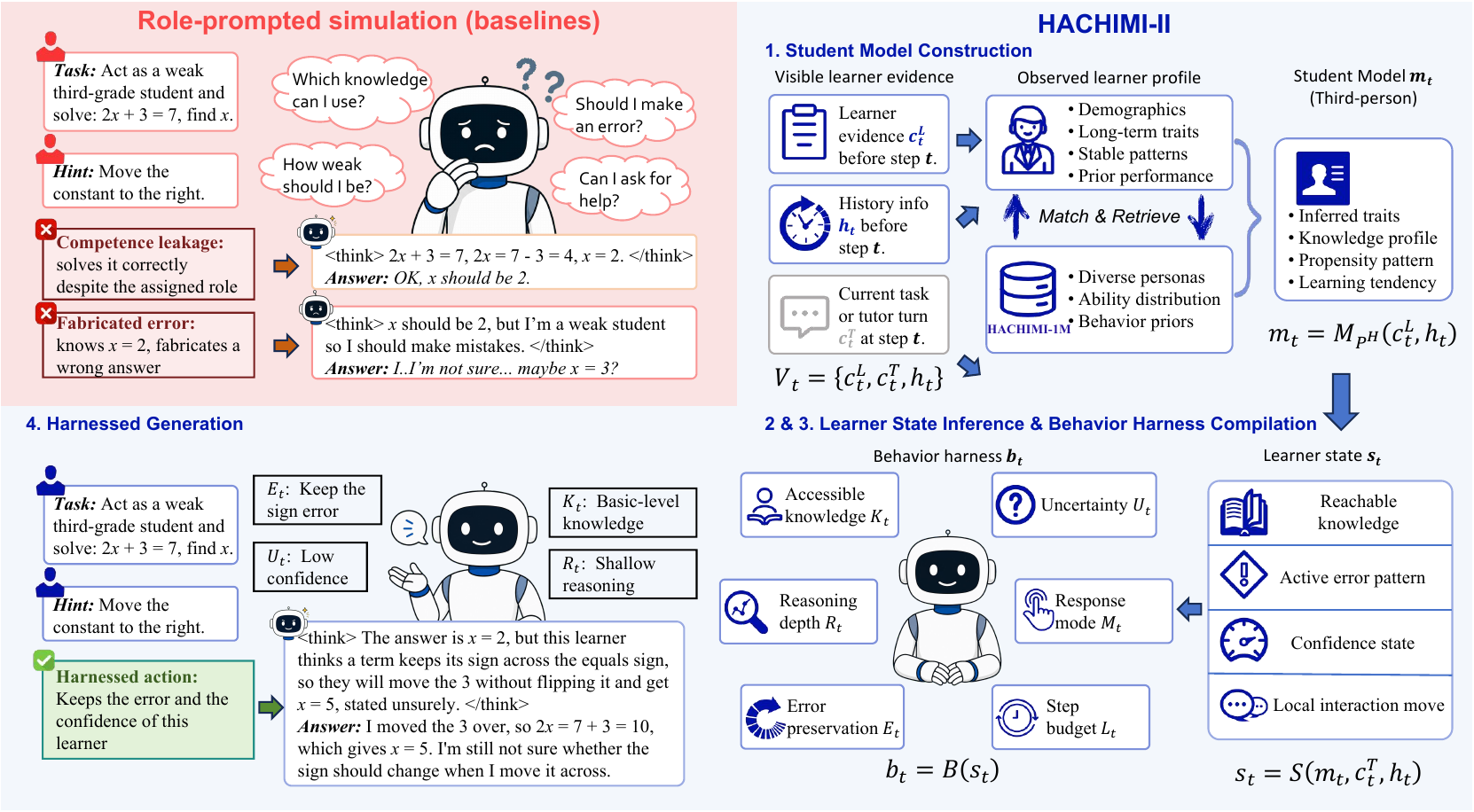}
\caption{Role-prompted simulation (top left) and the four steps of \methodname{} on a linear equation.}
\label{fig:overview}
\end{figure*}

\section{Introduction}

Simulated students let researchers evaluate educational content and compare teaching strategies before any real learner is involved, and they support the development of tutoring systems when real interaction data is scarce \citep{kaser2024simulated,scarlatos2026simulated}. To serve these uses, a simulated student has to behave like the learner it stands for. It should answer some items correctly, miss others, and make the errors that the real learner makes.

Large language models can write student answers and dialogue turns, and the usual way to obtain them is role prompting, which asks the model to act as a student of a given level \citep{benedetto2024using,scarlatos2026simulated}. However, a role prompt names a level without modeling what this learner knows or how the learner will respond to the current item or tutor turn. An aligned model is trained to be correct and helpful \citep{ouyang2022training,bai2022training} and already knows the content that the student is supposed to lack, so it fills the gap with its own knowledge and a generic picture of a student. As Figure~\ref{fig:overview} illustrates, it then either solves the problem despite the assigned role or inserts an error at a step where the learner would not fail. This failure has been described as the competence paradox \citep{yuan2026towards} and as expert knowledge leakage \citep{song2026novice}, and we call its observable effect \emph{competence leakage}. Under leakage, the ability level is set by the model, differences among learners largely disappear, and real errors are rarely reproduced.

Existing remedies add learner profiles \citep{lu2024generative}, agent memory \citep{xu2024eduagent}, or prompt-level constraints on what the student knows. They enrich the description of the student, but the model is still asked to act as the student and still has to suppress what it knows. Evaluations of prompting-based simulators report limited fidelity to real student behavior \citep{scarlatos2026simulated}.

In this paper, we counter competence leakage by modeling the learner's behavior, which changes the task framing from role-play to prediction. A request to act as a weak student asks the model to suppress knowledge it has. A request to predict what a particular learner will do next asks the model to use that knowledge, because the model must know the correct solution to judge where this learner will depart from it. Following this framing, we propose \methodname{} (Figure~\ref{fig:overview}). It describes the learner in the third person through a student model built from observed evidence, infers the learner's state for the current item or tutor turn, compiles that state into a behavior harness, and generates the next action within the bounds that the harness sets on knowledge, reasoning depth, errors, and commitment. \methodname{} takes its name from HACHIMI \citep{jiang2026hachimi}, whose corpus of one million student personas provides its learner prior, and adds the state inference and harness stages that turn a persona into a predicted action.

We evaluate \methodname{} against role-prompted baselines on five datasets in two formats. In item-response simulation on XES3G5M, Eedi, and ENEM, the Wasserstein-1 distance between the simulated and real distributions of per-student accuracy is 2.4--8.4$\times$ smaller than that of the strongest baseline, and \methodname{} recovers the ranking of individual students. In dialogue simulation on Eedi QATD2k and Khan/CoMTA, it reproduces real errors with the same knowledge type, locus, and mechanism 1.8--5.1$\times$ more often than the strongest baseline and more than doubles the F1 of error-state prediction. It also attains the highest agreement with the real learners on all five datasets.

This paper makes three contributions. First, we show that competence leakage in role-prompted simulation affects the ability level, the differences among learners, and their errors, and we recast the task as third-person prediction of the learner's next action. Second, we propose \methodname{}, which models the learner's behavior with a student model, a learner state inferred at each step, and a behavior harness compiled deterministically from that state. Third, on three item-response and two dialogue datasets, ablations show that the three components govern different parts of the simulated behavior, with the learner state as the main source of the gains.

\section{Related Work}

\paragraph{Learner modeling and student simulation.}
Real learner traces are costly to collect \citep{choi2020ednet,scarlatos2026simulated}. Early simulators used hand-specified cognitive models such as models of human memory \citep{reddy2017accelerating}, and learner models estimate proficiency through probabilistic mastery \citep{corbett1994knowledge}, item response theory \citep{lord1980applications}, or recurrent, memory-augmented, and reinforcement-learning architectures \citep{piech2015deep,zhang2017dynamic,zhao2023sim}. These models output correctness or mastery. The open-ended answers and dialogue turns that tutoring research needs are generated with language models.

\paragraph{LLM-based simulated students.}
Role prompting asks a language model to answer as a student of a given level \citep{benedetto2024using,scarlatos2026simulated}, profile-conditioned methods add knowledge states, misconceptions, histories, or personas \citep{lu2024generative,jiang2026hachimi}, and agent-based simulators add memory, reflection, and action modules \citep{xu2024eduagent,gao2025agent4edu}. \citet{marquezcarpintero2026review} review these architectures. Training-based simulators fine-tune a model on student data \citep{qi2026roleplaying} or specialize a pooled simulator to each student \citep{yang2026studentsim}, and \citet{song2026novice} unlearn knowledge so that the model answers like a novice. Simulated students, including fine-tuned ones, still show limited behavioral fidelity beyond answer accuracy \citep{scarlatos2026simulated,xu2026unveiling}. \methodname{} needs no training and keeps the model's knowledge, which it uses to predict where the learner will fail.

Three lines of work are closest to ours. \citet{wu2025embracing} build a cognitive prototype of each student from past learning records and refine a simulated solution toward the predicted performance. \citet{yuan2026towards} name the competence paradox and argue that valid simulation requires an explicit specification of the student's epistemic state. \citet{meng2026disengaged} condition simulated students on coarse engagement states. \methodname{} infers a finer state at every item and tutor turn, covering the predicted error, commitment, and help-seeking, and compiles it into constraints on the next action. Beyond answer-level ability alignment \citep{srivatsa2025can} and teacher judgments of realism \citep{martynova2025can}, we also compare the errors themselves in knowledge type, locus, and mechanism.

\paragraph{Constraints and harnesses for generation.}
In agent systems, a harness is an execution layer outside the model that organizes context, tools, memory, and verification \citep{li2026agentharness,pan2026natural,zhong2026ai}. Constrained generation and language-model programming shape outputs through explicit constraints and control flow \citep{garbacea2025challenging,beurer2023prompting}. \methodname{} applies this idea to behavior simulation. Its constraints are derived from a predicted learner state, and they restrict both the content and the form of the next action.

\section{\methodname}
\label{sec:method}

\methodname{} models the learner's behavior by predicting the next observable action in the four steps shown in Figure~\ref{fig:overview}. The same design serves item-response and dialogue simulation, with state and harness fields specific to each format (Appendix~\ref{app:fields}).

\subsection{Problem Formulation}
\label{sec:formulation}

At step $t$ the simulator observes $\mathcal{V}_t=\{c_t^L,c_t^T,h_t\}$, where $c_t^L$ is the learner evidence available before the step, $c_t^T$ is the current item or tutor turn, and $h_t$ is the visible history. The target action $a_t$ is the learner's response to one item in item-response simulation and the learner's next turn in dialogue simulation. Generation uses only $\mathcal{V}_t$. \methodname{} computes
\begin{equation}
\begin{alignedat}{2}
m_t &= \mathcal{M}_{\mathcal{P}^{\mathrm{H}}}(c_t^L,h_t), \quad & s_t &= \mathcal{S}(m_t,c_t^T,h_t),\\
b_t &= \mathcal{B}(s_t), \quad & a_t &= \mathcal{G}(\mathcal{V}_t,m_t,s_t,b_t),
\end{alignedat}
\end{equation}
where $m_t$ is the student model, $\mathcal{P}^{\mathrm{H}}$ is the HACHIMI-1M persona corpus \citep{jiang2026hachimi}, $s_t$ is the learner state, $b_t$ is the behavior harness, and $a_t$ is the generated action. $\mathcal{M}$, $\mathcal{S}$, and $\mathcal{G}$ are prompted calls to the backbone model, and $\mathcal{B}$ is a deterministic mapping.

\subsection{Student Model}

The student model $m_t$ describes who the learner is. It combines an observed profile, summarized from the learner's visible history, with a HACHIMI-1M persona that matches this profile (Appendix~\ref{app:persona}). The observed profile is the main evidence about the learner's ability and habits, and the persona adds a prior on longer-term background such as motivation, confidence, and interaction style. The student model is written in the third person, so the backbone reasons about the learner.

\subsection{Learner State Inference}

The learner state $s_t$ predicts how the learner will handle the current item or tutor turn. The state has four variables, shown in Figure~\ref{fig:overview} and Table~\ref{tab:state-harness}, and each follows an established line of educational research. Following evidence-centered design \citep{mislevy2003brief,mislevy2005evidence}, the reachable knowledge lists the prerequisites of the current item that the learner has mastered. Following the zone of proximal development \citep{vygotsky1978mind,wood1976role}, it also records how far the learner's reasoning can reach and how much of the tutor's support the learner takes up. Error analysis locates failures in the solution process and traces them to their source \citep{newman1977analysis,radatz1979error}, and the active error pattern accordingly predicts whether the learner will err and, if so, at which step and through which mechanism. Following research on self-regulated learning and help-seeking \citep{zimmerman2002becoming,panadero2017review,aleven2016help}, the confidence state records how firmly the learner commits to an answer, and the local interaction move records whether the learner attempts the step or asks for help.

\begin{table}[t]
\centering
\small
\setlength{\tabcolsep}{3pt}
\begin{tabularx}{\columnwidth}{@{}>{\raggedright\arraybackslash}p{0.34\columnwidth}>{\raggedright\arraybackslash}X@{}}
\toprule
State variable & Harness constraint on the next action \\
\midrule
Reachable knowledge & Accessible knowledge $K_t$ and reasoning depth $R_t$ of the action \\
Active error pattern & Error preservation $E_t$ at the predicted step \\
Confidence state & Uncertainty $U_t$ of the answer \\
Local interaction move & Response mode $M_t$ and step budget $L_t$ of the turn \\
\bottomrule
\end{tabularx}
\caption{Learner state variables and the harness constraints compiled from them, named as in Figure~\ref{fig:overview}.}
\label{tab:state-harness}
\end{table}

\subsection{Behavior Harness Compilation}

The compiler $\mathcal{B}$ turns the learner state into explicit constraints on the next action. It is a fixed mapping, so the same state always yields the same harness, and the generator receives concrete limits on what it may write. For example, an explanation depth of \emph{none} compiles to a limit of one sentence without a full derivation, and a predicted error at the process-skills step compiles to a requirement that the answer contain one concrete error at that step. Table~\ref{tab:state-harness} summarizes the mapping, and Appendix~\ref{app:fields} lists the harness fields for each format.

\subsection{Harnessed Generation}

The generator $\mathcal{G}$ receives the visible context, the student model, the state, and the harness, and writes the action. The generated action is checked against the harness and regenerated when it violates a constraint (Appendices~\ref{app:generation} and~\ref{app:prompts}).

Table~\ref{tab:example} follows one XES3G5M item through the four steps. The state predicts that the learner can compute the travel times but will average the two speeds, and the generated answer carries this error. The real learner answered this item incorrectly, while all three baselines solved it.

\begin{table}[t]
\centering
\small
\renewcommand{\arraystretch}{1.05}
\begin{tabularx}{\columnwidth}{@{}>{\raggedright\arraybackslash}p{0.21\columnwidth}>{\raggedright\arraybackslash}X@{}}
\toprule
\multicolumn{2}{@{}p{0.97\columnwidth}@{}}{\emph{Item.} Two towns are 300\,m apart. A student walks there at 10\,m/min and back at 15\,m/min. What is the average speed of the round trip in m/min? (Answer: 12)} \\
\midrule
Evidence & 30 earlier items with accuracy 0.80, and 0.43 on word problems; accuracy drops on harder items \\
Student model & Grade 3; confident on routine arithmetic; low confidence and reliance on help on unfamiliar problems \\
Learner state & Prerequisites familiar; item partly within reach; active error pattern: averages the two speeds, $(10+15)/2=12.5$; low commitment; short explanation \\
Harness & One final answer after a single reasoning step, in at most three sentences \\
Action & Reasoning step: 30 minutes there, 20 minutes back, average of 10 and 15. Answer: 12.5 \\
\midrule
Outcome & The real learner answered incorrectly. Direct RP, Prompt-C, and Profile-C all answer 12. \\
\bottomrule
\end{tabularx}
\caption{One XES3G5M item traced through the four steps of \methodname{}.}
\label{tab:example}
\end{table}

\section{Experimental Setup}
\label{sec:setup}

\begin{figure*}[t]
\centering
\includegraphics[width=\textwidth]{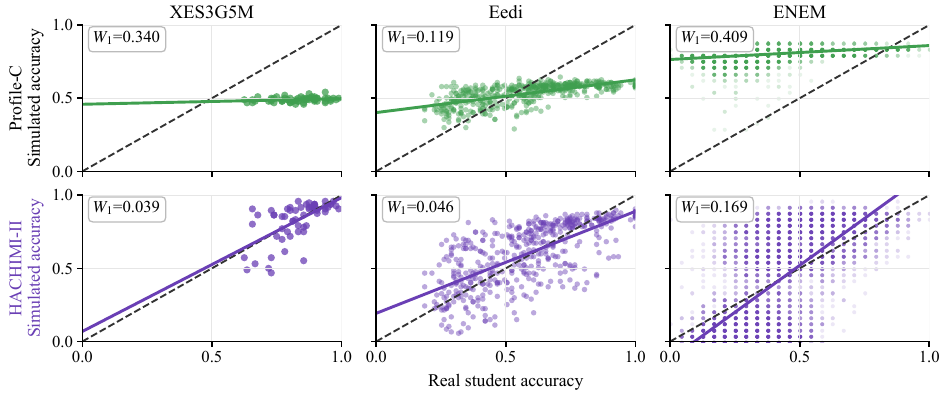}
\caption{Real and simulated accuracy per student for Profile-C and \methodname{}. The dashed line is $y=x$, and the solid line is a linear fit.}
\label{fig:static}
\end{figure*}

\paragraph{Datasets.}
Table~\ref{tab:datasets} lists the five datasets. The item-response datasets pair real students with items and record whether each response is correct. XES3G5M contains third-grade mathematics responses in Chinese, Eedi contains multiple-choice diagnostic questions in mathematics, and ENEM 2024 is the natural-sciences section of the Brazilian national secondary-school exam. The two dialogue datasets contain real mathematics tutoring sessions, with human tutors in Eedi QATD2k and an LLM tutor in Khan/CoMTA, and each instance is one student turn predicted from the preceding dialogue.

\begin{table}[t]
\centering
\small
\setlength{\tabcolsep}{3pt}
\begin{tabularx}{\columnwidth}{@{}l>{\raggedright\arraybackslash}X r@{}}
\toprule
Format & Dataset & Instances \\
\midrule
Item response & XES3G5M \citep{liu2023xes3g5m} & 60 $\times$ 96 \\
Item response & Eedi \citep{wang2021results} & 480 $\times$ 107 \\
Item response & ENEM 2024 Natural Sciences \citepalias{inep2024enemProvas} & 7{,}200 $\times$ 24 \\
Dialogue & Eedi QATD2k \citep{zent2025piivot} & 17{,}256 \\
Dialogue & Khan/CoMTA \citep{miller2024comta} & 917 \\
\bottomrule
\end{tabularx}
\caption{Datasets. Item-response instances are students $\times$ items, and dialogue instances are student turns.}
\label{tab:datasets}
\end{table}

\paragraph{Item-response metrics.}
For each student--item pair $n$, the real label $y_n\in\{0,1\}$ records whether the student answered correctly, and the simulated label $\hat{y}_n$ is obtained by matching the normalized simulated answer against the answer key. Let $\mathrm{acc}_i$ and $\widehat{\mathrm{acc}}_i$ be the real and simulated accuracy of student $i$ over the evaluated items. We report response agreement $\mathrm{Agree}=\frac{1}{N}\sum_n\mathbf{1}[y_n=\hat{y}_n]$ over the $N$ pairs, the Spearman correlation $\rho$ between $\mathrm{acc}_i$ and $\widehat{\mathrm{acc}}_i$, and the Wasserstein-1 distance \citep{villani2009optimal} between their distributions over the $S$ students,
\begin{equation}
W_1=\frac{1}{S}\sum_{k=1}^{S}\bigl|\mathrm{acc}_{(k)}-\widehat{\mathrm{acc}}_{(k)}\bigr|,
\end{equation}
where $\mathrm{acc}_{(k)}$ and $\widehat{\mathrm{acc}}_{(k)}$ are the $k$-th smallest real and simulated accuracies. $W_1$ measures whether the simulated population has the right ability level and spread, and $\rho$ measures whether it orders individual students correctly. Appendix~\ref{app:full} reports further metrics.

\begin{table}[t]
\centering
\small
\setlength{\tabcolsep}{4pt}
\begin{tabular}{llccc}
\toprule
Dataset & Method & Agree$\uparrow$ & $W_1\downarrow$ & $\rho\uparrow$ \\
\midrule
\multirow{4}{*}{XES3G5M}
& Direct RP & 0.534 & 0.327 & 0.006 \\
& Prompt-C & 0.533 & 0.332 & $-$0.071 \\
& Profile-C & 0.526 & 0.340 & 0.220 \\
& \methodname & \textbf{0.752} & \textbf{0.039} & \textbf{0.691} \\
\midrule
\multirow{4}{*}{Eedi}
& Direct RP & 0.502 & 0.160 & $-$0.057 \\
& Prompt-C & 0.509 & 0.155 & 0.311 \\
& Profile-C & 0.521 & 0.119 & 0.698 \\
& \methodname & \textbf{0.606} & \textbf{0.046} & \textbf{0.703} \\
\midrule
\multirow{4}{*}{ENEM}
& Direct RP & 0.473 & 0.432 & $-$0.008 \\
& Prompt-C & 0.475 & 0.426 & 0.303 \\
& Profile-C & 0.483 & 0.409 & 0.473 \\
& \methodname & \textbf{0.630} & \textbf{0.169} & \textbf{0.668} \\
\bottomrule
\end{tabular}
\caption{Item-response results. Best per dataset in bold.}
\label{tab:static}
\end{table}

\begin{table}[t]
\centering
\small
\setlength{\tabcolsep}{3pt}
\begin{tabularx}{\columnwidth}{@{}l>{\raggedright\arraybackslash}Xcc@{}}
\toprule
Dimension & Labels & Human & Judge \\
\midrule
Functional label & correct; incorrect; not applicable & 0.84 & 0.83 \\
\addlinespace
Error indicator & yes; no & 0.81 & 0.80 \\
\addlinespace
Knowledge type & factual; conceptual; procedural; metacognitive & 0.73 & 0.69 \\
\addlinespace
Error locus & reading; comprehension; transformation; process skills; encoding & 0.68 & 0.62 \\
\addlinespace
Error mechanism & language difficulty; spatial information difficulty; deficient prerequisites; incorrect associations; irrelevant rules & 0.66 & 0.60 \\
\bottomrule
\end{tabularx}
\caption{Judge labels and agreement. Human is Fleiss' $\kappa$ among three annotators, and Judge is Cohen's $\kappa$ between GPT-5.4 and their majority label.}
\label{tab:judge}
\end{table}

\paragraph{Dialogue metrics.}
A judge model, GPT-5.4, labels each real and simulated turn with the five dimensions in Table~\ref{tab:judge}. It assigns a functional label $f_n\in\{\text{correct},\text{incorrect},\text{NA}\}$ and an error indicator $e_n\in\{0,1\}$ that marks a mathematical error classifiable in our taxonomy. For each error it assigns three structure labels $\ell_n=(\ell_n^{\mathrm{type}},\ell_n^{\mathrm{locus}},\ell_n^{\mathrm{mech}})$, namely the knowledge type from the revised Bloom taxonomy \citep{anderson2001taxonomy}, the error locus from Newman's error analysis \citep{newman1977analysis}, and the error mechanism from Radatz's classification \citep{radatz1979error}. The judge was selected from five candidate models by its agreement with three human annotators on 200 turns per dimension, and its agreement with their majority label is within 0.06 of the agreement among the annotators on every dimension (Table~\ref{tab:judge}, Appendix~\ref{app:eval}).

We score three aspects of the simulated turns. Functional agreement uses Cohen's $\kappa_{\mathrm{func}}$ and the multiclass $\mathrm{MCC}_{\mathrm{func}}$ over $f_n$. Error-state prediction uses the balanced accuracy $\mathrm{BAcc}_{\mathrm{err}}$, $\mathrm{MCC}_{\mathrm{err}}$, and $\mathrm{F1}_{\mathrm{err}}$ over $e_n$. Error-structure fidelity counts a real error as reproduced only when the simulated turn also errs and all three structure labels match. With the real error set $\mathcal{E}=\{n:e_n=1\}$,
\begin{equation}
\mathrm{Rec}_{\mathrm{struct}}=\frac{1}{|\mathcal{E}|}\sum_{n\in\mathcal{E}}\mathbf{1}[\hat{e}_n=1\wedge\hat{\ell}_n=\ell_n]
\end{equation}
is the share of real errors that the simulator reproduces with matching structure. $\mathrm{Prec}_{\mathrm{struct}}$ is the corresponding share of simulated errors, and $\mathrm{F1}_{\mathrm{struct}}$ is their harmonic mean. Appendix~\ref{app:eval} gives all definitions.

\paragraph{Baselines.}
We compare \methodname{} with task-adapted versions of four simulation strategies. Direct Role-Play (Direct RP) asks the model to respond as a student given only the current item or the dialogue so far \citep{scarlatos2026simulated}. Prompt-Constrained Simulation (Prompt-C) adds natural-language constraints on accessible knowledge, reasoning depth, uncertainty, and error preservation. It is a prompt-level instantiation of the epistemic-state specification of \citet{yuan2026towards}. Profile-Conditioned Simulation (Profile-C) conditions on the learner's history and a learner profile \citep{lu2024generative,liu2024personality,duan2026who,jiang2026hachimi}, which in item-response simulation is the student model of \methodname{}. Agent-Style Memory Simulation (Agent-Mem), used in dialogue only, adds a learner profile and memory of the visible dialogue \citep{xu2024eduagent,gao2025agent4edu}. All methods use the same backbone, Qwen3.5-397B, with temperature 0.2 (Appendix~\ref{app:impl}).

\paragraph{Ablations.}
\emph{w/o Prior} builds the student model from visible evidence only, without a HACHIMI-1M persona. \emph{w/o State} removes learner state inference. The generator receives only the student model and the task context, and the harness keeps only its output-format constraints. \emph{w/o Harness} gives the generator the student model and the inferred state without harness compilation.

\section{Results}
\label{sec:results}

\subsection{Item-Response Simulation}

Table~\ref{tab:static} shows that \methodname{} attains the highest agreement, the smallest $W_1$, and the highest $\rho$ on all three datasets. It gives the same outcome as the real learner on 75\%, 61\%, and 63\% of the student--item pairs, against at most 53\%, 52\%, and 48\% for the baselines. Its $W_1$ is 2.4--8.4$\times$ smaller than that of the strongest baseline on each dataset.

The baselines exhibit competence leakage at the population level. Each baseline simulates one generic student and assigns nearly the same accuracy to every learner, at a level set by the model and the prompt. This level sits near 0.50 on XES3G5M, where the real mean is 0.83, and near 0.82 on ENEM, where the real mean is 0.39. Direct RP sees no learner history, and its $\rho$ stays between $-0.06$ and $0.01$. With the student model, Profile-C recovers the ordering of students on Eedi ($\rho=0.698$), yet its simulated accuracies still cluster around one value ($W_1=0.119$).

\methodname{} tracks both the level and the spread of real ability. Its simulated mean accuracy is 0.828, 0.593, and 0.384 on the three datasets against real means of 0.831, 0.574, and 0.394 (Appendix~\ref{app:ablation}). In Figure~\ref{fig:static}, the Profile-C points stay within a narrow band, while the \methodname{} points follow the diagonal, and $\rho$ of 0.67--0.70 shows that the method also recovers the ordering of individual learners.

Table~\ref{tab:bands} makes the contrast concrete by splitting the students of each dataset into thirds by real accuracy. The baselines give the three groups almost the same simulated accuracy. On ENEM, the weakest third answers 22\% of the items correctly, and the baselines simulate these students at 78--83\%. \methodname{} follows the real level of every group. It simulates the three ENEM groups at 15\%, 39\%, and 68\% against real values of 22\%, 39\%, and 61\%. The strongest and the weakest third differ by at most 0.11 for any baseline, against 0.19--0.47 for the real students and 0.18--0.53 for \methodname{}.

\begin{table*}[t]
\centering
\small
\setlength{\tabcolsep}{6pt}
\begin{tabular}{lccccccccc}
\toprule
& \multicolumn{3}{c}{XES3G5M} & \multicolumn{3}{c}{Eedi} & \multicolumn{3}{c}{ENEM} \\
\cmidrule(lr){2-4}\cmidrule(lr){5-7}\cmidrule(lr){8-10}
& Low & Middle & High & Low & Middle & High & Low & Middle & High \\
\midrule
Real students & 0.74 & 0.85 & 0.93 & 0.34 & 0.58 & 0.81 & 0.22 & 0.39 & 0.61 \\
\midrule
Direct RP & 0.50 & 0.50 & 0.50 & 0.59 & 0.59 & 0.59 & 0.83 & 0.83 & 0.83 \\
Prompt-C & 0.50 & 0.50 & 0.50 & 0.58 & 0.59 & 0.60 & 0.81 & 0.82 & 0.83 \\
Profile-C & 0.49 & 0.49 & 0.49 & 0.47 & 0.54 & 0.58 & 0.78 & 0.80 & 0.82 \\
\methodname & 0.74 & 0.84 & 0.92 & 0.42 & 0.61 & 0.75 & 0.15 & 0.39 & 0.68 \\
\bottomrule
\end{tabular}
\caption{Mean per-student accuracy by third of real accuracy.}
\label{tab:bands}
\end{table*}

\subsection{Dialogue Simulation}
\label{sec:dialogue}

\begin{table*}[t]
\centering
\small
\setlength{\tabcolsep}{5pt}
\begin{tabular}{llccccccc}
\toprule
& & \multicolumn{2}{c}{Functional agreement} & \multicolumn{3}{c}{Error-state prediction} & \multicolumn{2}{c}{Error-structure fidelity} \\
\cmidrule(lr){3-4}\cmidrule(lr){5-7}\cmidrule(lr){8-9}
Dataset & Method & $\kappa_{\mathrm{func}}\uparrow$ & $\mathrm{MCC}_{\mathrm{func}}\uparrow$ & $\mathrm{BAcc}_{\mathrm{err}}\uparrow$ & $\mathrm{MCC}_{\mathrm{err}}\uparrow$ & $\mathrm{F1}_{\mathrm{err}}\uparrow$ & $\mathrm{Rec}_{\mathrm{struct}}\uparrow$ & $\mathrm{F1}_{\mathrm{struct}}\uparrow$ \\
\midrule
\multirow{5}{*}{Eedi QATD2k}
& Direct RP & 0.592 & 0.616 & 0.560 & 0.175 & 0.236 & 0.157 & 0.220 \\
& Prompt-C & 0.586 & 0.614 & 0.558 & 0.174 & 0.230 & 0.150 & 0.214 \\
& Profile-C & 0.595 & 0.615 & 0.565 & 0.179 & 0.248 & 0.172 & 0.232 \\
& Agent-Mem & 0.597 & 0.618 & 0.562 & 0.177 & 0.240 & 0.162 & 0.225 \\
& \methodname & \textbf{0.659} & \textbf{0.673} & \textbf{0.738} & \textbf{0.509} & \textbf{0.561} & \textbf{0.315} & \textbf{0.339} \\
\midrule
\multirow{5}{*}{Khan/CoMTA}
& Direct RP & 0.273 & 0.337 & 0.515 & 0.052 & 0.134 & 0.080 & 0.127 \\
& Prompt-C & 0.274 & 0.345 & 0.512 & 0.045 & 0.117 & 0.067 & 0.110 \\
& Profile-C & 0.289 & 0.347 & 0.530 & 0.098 & 0.183 & 0.110 & 0.169 \\
& Agent-Mem & 0.288 & 0.354 & 0.531 & 0.110 & 0.175 & 0.105 & 0.168 \\
& \methodname & \textbf{0.370} & \textbf{0.390} & \textbf{0.660} & \textbf{0.267} & \textbf{0.464} & \textbf{0.565} & \textbf{0.346} \\
\bottomrule
\end{tabular}
\caption{Dialogue results. Best per dataset in bold.}
\label{tab:dynamic}
\end{table*}

\begin{table*}[t]
\centering
\small
\setlength{\tabcolsep}{4.2pt}
\begin{tabular}{lcccccccccc}
\toprule
& \multicolumn{3}{c}{$W_1\downarrow$} & \multicolumn{3}{c}{$\rho\uparrow$} & \multicolumn{2}{c}{$\kappa_{\mathrm{func}}\uparrow$} & \multicolumn{2}{c}{$\mathrm{F1}_{\mathrm{struct}}\uparrow$} \\
\cmidrule(lr){2-4}\cmidrule(lr){5-7}\cmidrule(lr){8-9}\cmidrule(lr){10-11}
Variant & XES & Eedi & ENEM & XES & Eedi & ENEM & QATD2k & CoMTA & QATD2k & CoMTA \\
\midrule
\methodname & 0.039 & 0.046 & 0.169 & 0.691 & 0.703 & 0.668 & 0.659 & 0.370 & 0.339 & 0.346 \\
w/o Prior & 0.025 & 0.041 & 0.142 & 0.603 & 0.704 & 0.594 & 0.571 & 0.326 & 0.264 & 0.323 \\
w/o State & 0.074 & 0.116 & 0.409 & 0.664 & 0.742 & 0.539 & 0.741 & 0.382 & 0.136 & 0.119 \\
w/o Harness & 0.056 & 0.048 & 0.257 & 0.559 & 0.703 & 0.643 & 0.565 & 0.348 & 0.290 & 0.346 \\
\bottomrule
\end{tabular}
\caption{Ablation results. Appendix~\ref{app:ablation} reports the remaining metrics.}
\label{tab:ablation}
\end{table*}

\begin{figure}[t]
\centering
\includegraphics[width=\columnwidth]{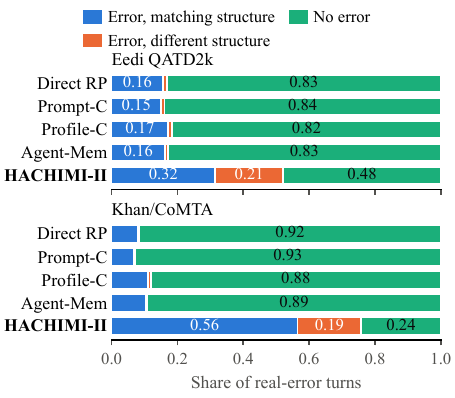}
\caption{Simulated outputs on turns where the real learner errs.}
\label{fig:destination}
\end{figure}

Table~\ref{tab:dynamic} reports the dialogue results. Functional agreement is dominated by the frequent turns that give a correct answer or contain no mathematics, so the methods separate most clearly on the turns where the learner errs. Figure~\ref{fig:destination} shows what each simulator produces on these turns. The baselines answer 82--93\% of them without any error, which is competence leakage at the level of a single turn, while \methodname{} errs on 52\% of them on Eedi QATD2k and 76\% on Khan/CoMTA.

\methodname{} predicts when the learner errs and reproduces the error. Its $\mathrm{F1}_{\mathrm{err}}$ of 0.561 on Eedi QATD2k and 0.464 on Khan/CoMTA is 2.3$\times$ and 2.5$\times$ that of the strongest baseline. It reproduces 31.5\% and 56.5\% of real errors with all three structure labels matching, which is 1.8$\times$ and 5.1$\times$ the rate of the strongest baseline, and its $\mathrm{F1}_{\mathrm{struct}}$ is 1.5$\times$ and 2.0$\times$ that of the strongest baseline. \methodname{} also attains the highest $\kappa_{\mathrm{func}}$ and $\mathrm{MCC}_{\mathrm{func}}$ on both datasets, so its gains on error turns come with higher functional agreement than every baseline.

Table~\ref{tab:case} shows one Eedi QATD2k turn. The real learner fails to distribute the 3 when expanding $3(n-4)$, and all four baselines return the correct expansion. \methodname{} predicts this error in the learner state and writes $3n-4$, which the judge labels with the same knowledge type, locus, and mechanism as the real turn.

\subsection{What Each Component Contributes}
\label{sec:ablation}

Table~\ref{tab:ablation} removes one component at a time. The three components control different parts of the simulated behavior.

\paragraph{The learner state counters competence leakage.}
Without the state, the simulated population moves toward the model's own competence. On ENEM the simulated mean accuracy rises from 0.384 to 0.802 against a real 0.394, and $W_1$ grows 1.9--2.5$\times$ across the three item-response datasets, returning to the level of Profile-C on Eedi and ENEM. In dialogue, $\mathrm{F1}_{\mathrm{struct}}$ falls from 0.339 to 0.136 on Eedi QATD2k and from 0.346 to 0.119 on Khan/CoMTA. This variant still builds the full student model, including the persona, and still generates through separate model calls, so these gains come from the explicit prediction of the learner state.

The same variant shows that functional agreement can be high under leakage. Its $\kappa_{\mathrm{func}}$ is 0.741 on Eedi QATD2k because it matches the frequent correct and non-mathematical turns, while it reproduces 9.9\% of real errors with matching structure.

\paragraph{The harness keeps the action consistent with the learner.}
Removing the harness increases $W_1$ from 0.039 to 0.056 on XES3G5M and from 0.169 to 0.257 on ENEM, and lowers $\kappa_{\mathrm{func}}$ on both dialogue datasets. On Eedi QATD2k the errors generated without the harness match the learner's error less often, and $\mathrm{F1}_{\mathrm{struct}}$ drops from 0.339 to 0.290. The harness places errors at the predicted step and keeps the rest of the turn in line with the learner.

\paragraph{The persona prior improves the ordering of learners.}
Removing the HACHIMI-1M prior lowers $\rho$ on XES3G5M from 0.691 to 0.603 and on ENEM from 0.668 to 0.594. On Eedi QATD2k it lowers $\kappa_{\mathrm{func}}$ from 0.659 to 0.571 and $\mathrm{F1}_{\mathrm{struct}}$ from 0.339 to 0.264, and both also fall on Khan/CoMTA. $W_1$ alone is lower without the prior, by 0.005 to 0.027. The persona sharpens the differences between weaker and stronger learners, which improves their ordering and, on XES3G5M and ENEM, also widens the spread of simulated accuracy (Appendix~\ref{app:ablation}). $W_1$ compares the two distributions as wholes, so it counts the wider spread and does not register the better ordering. On ENEM the prior also corrects the simulated mean, to 0.384 from 0.500 against a real 0.394.

\begin{table}[t]
\centering
\small
\renewcommand{\arraystretch}{1.05}
\begin{tabularx}{\columnwidth}{@{}>{\raggedright\arraybackslash}p{0.2\columnwidth}>{\raggedright\arraybackslash}X@{}}
\toprule
\multicolumn{2}{@{}p{0.97\columnwidth}@{}}{\emph{Question.} A function machine subtracts 4 and then multiplies by 3. Leo writes $3n-12$ and Sophie writes $3(n-4)$. Who is correct?} \\
\midrule
Tutor & What do you get if you multiply out the brackets? \\
Real learner & 3n-4 (incorrect; procedural, process skills, deficient prerequisites) \\
Baselines & $3n-12$ (all four; correct) \\
\methodname & \emph{State.} Weak self-monitoring, tentative, fragment only. \emph{Predicted error.} $3n-4$ in place of $3n-12$. \emph{Harness.} One concrete error at the process-skills step, one sentence. \emph{Turn.} i get 3n-4 (incorrect; same labels on all three dimensions) \\
\bottomrule
\end{tabularx}
\caption{A real Eedi QATD2k turn (\S\ref{sec:dialogue}).}
\label{tab:case}
\end{table}

\section{Discussion}
\label{sec:discussion}

Our results show that an explicit model of the learner's behavior counters competence leakage. Under role prompting, each simulator portrays one generic student whose accuracy is set by the model and the prompt. This uniformity is the ability-level counterpart of the flattening and caricature observed when language models simulate opinions, personas, and identity groups \citep{santurkar2023whose,cheng2023compost,wang2025large}. As in social simulation, where interactions scripted by one model look more successful than those among agents that hold only their own information \citep{zhou2024real}, the simulator draws on knowledge that the simulated party does not have. Prediction changes the role of the model's knowledge. To predict that a learner will average two speeds, the model must know how average speed is computed and where this learner departs from it, so the competence that leaks under role-play becomes the source of faithful errors. Evaluations of simulated students should compare the simulated ability distribution with the real one and score the error state and its structure together with functional agreement, which is dominated by turns without an error. A simulator that predicts how a particular learner responds to an item or a tutor move supports uses that depend on ability differences or on specific misconceptions, such as comparing teaching strategies across ability levels or testing a hint against a misconception. It is also a step toward a world model of the learner \citep{ha2018recurrent,zhou2025social}, against which tutoring policies can be planned and compared before they reach real students.

\section{Conclusion}

Role-prompted student simulators exhibit competence leakage and portray a generic student whose ability is set by the model. \methodname{} models the learner's behavior through third-person prediction, inferring the learner's state for the current item or tutor turn and compiling it into a behavior harness for the next action. Compared with every baseline, it brings the simulated ability distribution closer to the real one on three item-response datasets and reproduces real errors with their full structure more often on two dialogue datasets. Faithful student simulation thus rests on predicting the learner, which lets tutor development and evaluation use capable models as simulated students.

\section*{Limitations}

Our evaluation covers the correctness of item responses and the presence and structure of mathematical errors in tutoring turns, on datasets in mathematics and natural sciences. A learner's next action also depends on factors that no record captures, such as momentary attention and the outcome of a guess, and \methodname{} predicts the part of behavior that the available evidence determines. Each dialogue turn is predicted from the real preceding dialogue. A direct extension is multi-turn simulation in which the simulated student's own turns form the history, evaluated with a protocol that compares generated and real dialogues over many turns. The state and harness are defined through knowledge, reasoning, errors, confidence, and interaction, and the method uses the backbone through prompts and deterministic rules, so it can be applied to further subjects and backbones without training.

\section*{Ethical Considerations}

All datasets in this work are publicly available, and we use them under their original terms of use. The learner records in these datasets are de-identified, and \methodname{} uses only their responses, dialogue turns, and grade or age. A matched persona comes from HACHIMI-1M, a corpus of generated student personas \citep{jiang2026hachimi}, and is used only as a prior on learning behavior inside the simulator. The simulated students are intended to supplement data from real learners when educational systems are developed and evaluated. Each output is a predicted response to one item or tutor turn and should not be used to assess or classify individual students.

\bibliography{references}

\appendix
\section{Persona Matching}
\label{app:persona}

\methodname{} retrieves one HACHIMI-1M persona for each learner in six steps.
\begin{enumerate}
\item \textbf{Observed profile.} The learner's visible history is summarized into a profile with grade or age, ability level, past performance, strengths and weaknesses, response style, uncertainty, and help-seeking.
\item \textbf{Candidate retrieval.} A candidate pool is drawn from HACHIMI-1M on coarse fields, namely grade or age range, academic level, subject strengths and weaknesses, and learning-style tags.
\item \textbf{Rule-based ranking.} Candidates are scored on grade compatibility, agreement of ability level, overlap of strong and weak subjects, and similarity of learning behavior.
\item \textbf{Reranking.} The backbone compares the observed profile with the full descriptions of the top-$k$ candidates ($k=8$), using motivation, confidence, learning habits, self-efficacy, and interaction style.
\item \textbf{Validity check.} Near-duplicate candidates are removed, and the selected persona is checked for membership in the candidate pool and completeness of its fields.
\item \textbf{Fusion.} The selected persona is merged with the observed profile into the third-person student model.
\end{enumerate}
\section{State and Harness Fields}
\label{app:fields}

Table~\ref{tab:fields} lists the fields used in each format. Both formats follow the design in Table~\ref{tab:state-harness}, and the fields differ because the evidence and the target action differ. Prerequisite familiarity, reachability, and the uptake of the tutor's support refine the reachable knowledge. The likely error path in item-response simulation and the misconception status and predicted error in dialogue form the active error pattern. Commitment and self-monitoring form the confidence state, and help-seeking, readiness to attempt, the form of engagement, and response depth form the local interaction move. In item-response simulation the state describes the learner's relation to the current item, and the harness constrains the answer and its explanation. In dialogue the state describes the learner's response to the current tutor turn, and the harness constrains what the next turn may say and do.

\begin{table*}[t]
\centering
\small
\setlength{\tabcolsep}{4pt}
\renewcommand{\arraystretch}{1.15}
\begin{tabularx}{\textwidth}{@{}p{0.14\textwidth}>{\raggedright\arraybackslash}X>{\raggedright\arraybackslash}X@{}}
\toprule
Component & Item-response simulation & Dialogue simulation \\
\midrule
Input evidence
& Learner ability, prior responses, guessing tendency, confidence calibration; item prerequisites, operation complexity, response format, guessability
& Visible dialogue history, previous student turn, current tutor turn; type, strength, and target of the tutor's support \\
\midrule
Learner state
& Prerequisite familiarity, whether the item is within reach, likely error path, commitment, depth of explanation
& Whether the step is within reach, misconception status, self-monitoring, uptake of the tutor's support, help-seeking, readiness to attempt, form of engagement, commitment, depth of response, predicted error (presence, locus, mechanism, faulty rule) \\
\midrule
Behavior harness
& Whether a full explanation is allowed, response-length limit, restriction on fully correct reasoning, error path to follow, answer commitment
& Whether the answer may be stated directly, whether full reasoning is allowed, persistence of confusion and of the predicted misconception, whether help-seeking is allowed, number of solution steps \\
\bottomrule
\end{tabularx}
\caption{State and harness fields in each task format.}
\label{tab:fields}
\end{table*}

\section{Generation and Checking}
\label{app:generation}

In dialogue simulation, each generated turn is rewritten in the learner's register, and a consistency check confirms that its content is unchanged. A rule-based check and a review by the backbone then compare the turn with the harness. They look for a response more complete than the state allows, a transition faster than the predicted state allows, and a break in the predicted error path. A turn that violates the harness is regenerated once under tightened constraints. In item-response simulation, the output contains a final answer and a short student response, and it is checked and regenerated in the same way.

\section{Evaluation Details}
\label{app:eval}

\paragraph{Item-response metrics.}
Item-response evaluation extracts the final answer from the simulated response, normalizes whitespace, letter case, and option markers, and compares the result with the answer key to obtain $\hat{y}_n$. A $W_1$ of 0.10 corresponds to an average gap of ten accuracy points between real and simulated students of the same rank.

\paragraph{Judge labels.}
In dialogue, the judge labels the real and the simulated turn with the five dimensions in Table~\ref{tab:judge}. All metrics are computed from these labels.

\paragraph{Judge selection.}
We chose the judge by agreement with human annotation. For each of the five dimensions, three annotators labeled 200 turns, and the turns for the three structure dimensions all contain an error. The annotators are graduate students in AI and NLP who were recruited for this task and paid at or above the local standard rate, and they consented to the use of their labels in this study. The majority label is the gold standard. Table~\ref{tab:judge-irr} reports the agreement among annotators, and Table~\ref{tab:judge-selection} reports the agreement of five candidate judges with the gold labels. GPT-5.4 attains the highest mean agreement and the highest agreement on the functional label, the error indicator, and the knowledge type, and we use it for all dialogue evaluation. The judge and the simulation backbone come from different model families.

\begin{table}[t]
\centering
\small
\begin{tabular}{lcc}
\toprule
Dimension & Fleiss' $\kappa$ & Agreement (\%) \\
\midrule
Functional label & 0.84 & 91.0 \\
Error indicator & 0.81 & 90.5 \\
Knowledge type & 0.73 & 82.5 \\
Error locus & 0.68 & 78.5 \\
Error mechanism & 0.66 & 77.0 \\
\bottomrule
\end{tabular}
\caption{Agreement among the three annotators. Agreement (\%) is exact three-way agreement.}
\label{tab:judge-irr}
\end{table}

\begin{table}[t]
\centering
\small
\setlength{\tabcolsep}{3pt}
\begin{tabular}{lcccccc}
\toprule
Judge & Func. & Err. & Type & Locus & Mech. & Mean \\
\midrule
Gemini 3.5 & 0.79 & 0.74 & 0.64 & 0.58 & 0.55 & 0.660 \\
Kimi 2.6 & 0.76 & 0.72 & 0.62 & 0.56 & 0.54 & 0.640 \\
GLM 5.1 & 0.78 & 0.73 & 0.63 & 0.57 & 0.55 & 0.652 \\
DeepSeek V4 & 0.80 & 0.77 & 0.68 & \textbf{0.64} & \textbf{0.62} & 0.702 \\
GPT-5.4 & \textbf{0.83} & \textbf{0.80} & \textbf{0.69} & 0.62 & 0.60 & \textbf{0.708} \\
\bottomrule
\end{tabular}
\caption{Cohen's $\kappa$ between each candidate judge and the majority human labels.}
\label{tab:judge-selection}
\end{table}

\paragraph{Functional agreement.}
Let $C\in\mathbb{N}^{3\times3}$ be the confusion matrix of real and simulated functional labels, with row sums $t_i$, column sums $p_j$, and total $N$. Cohen's $\kappa=(p_o-p_e)/(1-p_e)$ uses the observed agreement $p_o=\frac{1}{N}\sum_iC_{ii}$ and the chance agreement $p_e=\frac{1}{N^2}\sum_kt_kp_k$. The multiclass Matthews correlation coefficient is
\begin{equation}
\mathrm{MCC}=\frac{N\sum_kC_{kk}-\sum_kt_kp_k}{\sqrt{\bigl(N^2-\sum_kp_k^2\bigr)\bigl(N^2-\sum_kt_k^2\bigr)}}.
\end{equation}
Both equal 1 for perfect agreement and approach 0 for a simulator whose labels are independent of the real ones.

\paragraph{Error-state prediction.}
We treat the error indicator as a binary target. With the real error set $\mathcal{E}$ and the simulated error set $\hat{\mathcal{E}}=\{n:\hat{e}_n=1\}$, the four counts are the true positives $\mathrm{TP}=|\mathcal{E}\cap\hat{\mathcal{E}}|$, the false positives $\mathrm{FP}=|\hat{\mathcal{E}}\setminus\mathcal{E}|$, the false negatives $\mathrm{FN}=|\mathcal{E}\setminus\hat{\mathcal{E}}|$, and the true negatives $\mathrm{TN}$. Error recall is $\mathrm{TP}/(\mathrm{TP}+\mathrm{FN})$, error precision is $\mathrm{TP}/(\mathrm{TP}+\mathrm{FP})$, and $\mathrm{F1}_{\mathrm{err}}$ is their harmonic mean. $\mathrm{BAcc}_{\mathrm{err}}$ is the mean of error recall and the specificity $\mathrm{TN}/(\mathrm{TN}+\mathrm{FP})$. $\mathrm{MCC}_{\mathrm{err}}$ and $\kappa_{\mathrm{err}}$ are computed from the same four counts, and both equal 0 for a simulator that always errs or never errs.

\paragraph{Error-structure fidelity.}
$\mathrm{Rec}_{\mathrm{struct}}$ is defined in \S\ref{sec:setup}, and
\begin{equation}
\mathrm{Prec}_{\mathrm{struct}}=\frac{1}{|\hat{\mathcal{E}}|}\sum_{n\in\hat{\mathcal{E}}}\mathbf{1}[e_n=1\wedge\hat{\ell}_n=\ell_n]
\end{equation}
is the share of simulated errors that match a real error in structure. A pair of turns counts as a structural match only when both contain an error and the knowledge type, locus, and mechanism all agree. Both metrics are bounded by their error-state counterparts, $\mathrm{Rec}_{\mathrm{struct}}\le$ error recall and $\mathrm{Prec}_{\mathrm{struct}}\le$ error precision.

\section{Implementation Details}
\label{app:impl}

All methods follow the protocol of \S\ref{sec:formulation}. In item-response simulation each method receives the current question and its answer options, and every method except Direct RP also receives the learner's history. The history contains 30 of the student's responses on XES3G5M and Eedi and 20 on ENEM, all to items outside the test set, and every student of a dataset answers the same test items. In dialogue each method receives the dialogue before the target turn. Appendix~\ref{app:prompts} gives the prompts of \methodname{}.

\paragraph{Direct Role-Play.}
The model acts as a student and generates the action from the current item or the dialogue so far.

\paragraph{Prompt-Constrained Simulation.}
Direct role-play is extended with natural-language instructions that constrain accessible knowledge, reasoning depth, uncertainty, and error preservation. All control stays in the prompt.

\paragraph{Profile-Conditioned Simulation.}
Generation is conditioned on the visible context and a learner profile. In item-response simulation the profile is the student model built by \methodname{}, and in dialogue it is a coarse profile computed from the learner's earlier turns.

\paragraph{Agent-Style Memory Simulation.}
Used only in dialogue, this baseline conditions on the visible dialogue, a coarse learner profile, and a short-term memory of the interaction.

\section{Full Results}
\label{app:full}

\begin{figure*}[t]
\centering
\includegraphics[width=\textwidth]{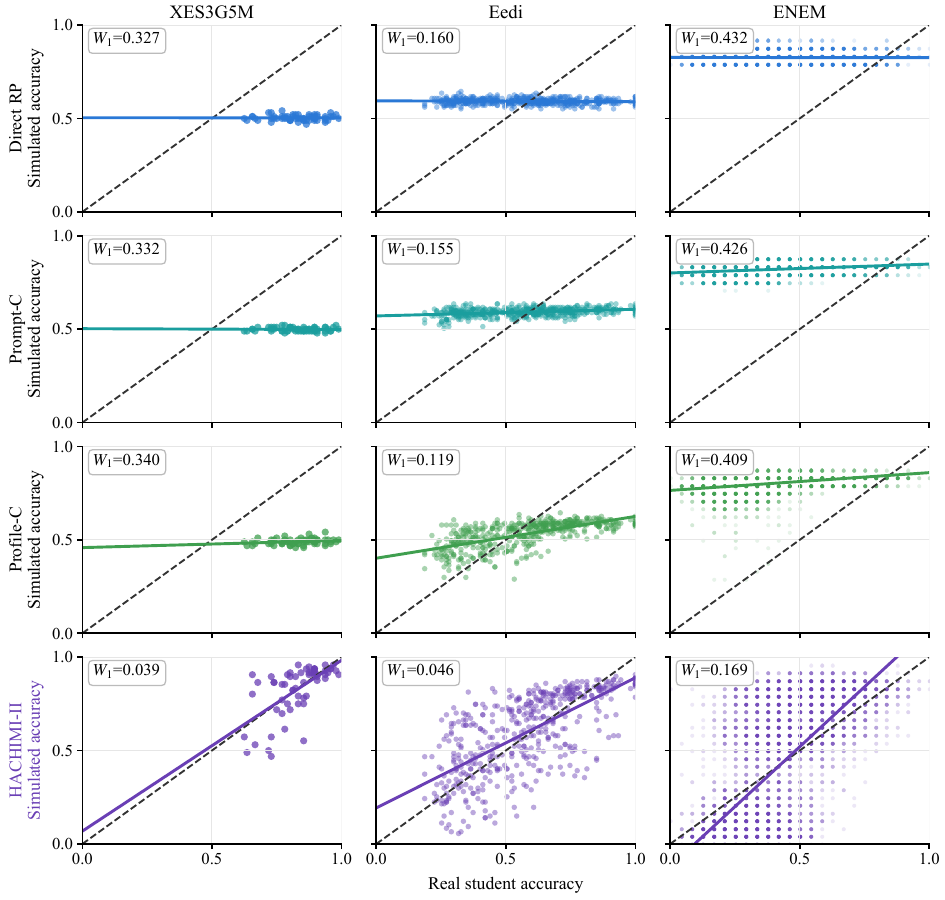}
\caption{Real and simulated accuracy per student for all methods and datasets. The dashed line is $y=x$, and the solid line is a linear fit.}
\label{fig:static-all}
\end{figure*}

Table~\ref{tab:static-supp} reports additional item-response metrics, namely the Kolmogorov--Smirnov statistic between the real and simulated distributions of per-student accuracy, the Pearson correlation of per-student accuracy, and the Spearman and Pearson correlations of per-item accuracy. On all three datasets, \methodname{} has the smallest Kolmogorov--Smirnov statistic and the highest Pearson correlations of per-student and per-item accuracy. Figure~\ref{fig:static-all} shows per-student accuracy for every method. Table~\ref{tab:dynamic-supp} reports additional dialogue metrics. On both datasets, \methodname{} has the highest error recall and $\kappa_{\mathrm{err}}$.

\begin{table}[t]
\centering
\footnotesize
\setlength{\tabcolsep}{1.5pt}
\begin{tabular}{llcccc}
\toprule
Dataset & Method & KS$\downarrow$ & Stu.\ $r\uparrow$ & Item $\rho\uparrow$ & Item $r\uparrow$ \\
\midrule
\multirow{4}{*}{XES3G5M}
& Direct RP & 1.000 & $-$0.005 & 0.364 & 0.260 \\
& Prompt-C & 1.000 & $-$0.040 & 0.366 & 0.295 \\
& Profile-C & 1.000 & 0.215 & 0.348 & 0.264 \\
& \methodname & 0.217 & 0.629 & 0.329 & 0.315 \\
\midrule
\multirow{4}{*}{Eedi}
& Direct RP & 0.448 & $-$0.062 & $-$0.128 & $-$0.082 \\
& Prompt-C & 0.415 & 0.334 & $-$0.085 & $-$0.044 \\
& Profile-C & 0.408 & 0.654 & $-$0.033 & $-$0.006 \\
& \methodname & 0.115 & 0.675 & 0.432 & 0.429 \\
\midrule
\multirow{4}{*}{ENEM}
& Direct RP & 0.973 & $-$0.009 & 0.511 & 0.416 \\
& Prompt-C & 0.956 & 0.290 & 0.461 & 0.399 \\
& Profile-C & 0.918 & 0.391 & 0.479 & 0.416 \\
& \methodname & 0.370 & 0.677 & 0.638 & 0.500 \\
\bottomrule
\end{tabular}
\caption{Additional item-response metrics.}
\label{tab:static-supp}
\end{table}

\begin{table*}[t]
\centering
\small
\setlength{\tabcolsep}{5pt}
\begin{tabular}{llcccc}
\toprule
Dataset & Method & Error recall & Error precision & $\kappa_{\mathrm{err}}$ & $\mathrm{Prec}_{\mathrm{struct}}$ \\
\midrule
\multirow{5}{*}{Eedi QATD2k}
& Direct RP & 0.168 & 0.393 & 0.156 & 0.368 \\
& Prompt-C & 0.161 & 0.404 & 0.153 & 0.375 \\
& Profile-C & 0.184 & 0.382 & 0.164 & 0.358 \\
& Agent-Mem & 0.173 & 0.392 & 0.160 & 0.368 \\
& \methodname & 0.521 & 0.608 & 0.507 & 0.368 \\
\midrule
\multirow{5}{*}{Khan/CoMTA}
& Direct RP & 0.084 & 0.333 & 0.039 & 0.316 \\
& Prompt-C & 0.071 & 0.327 & 0.032 & 0.306 \\
& Profile-C & 0.119 & 0.397 & 0.078 & 0.368 \\
& Agent-Mem & 0.110 & 0.431 & 0.082 & 0.414 \\
& \methodname & 0.758 & 0.335 & 0.220 & 0.249 \\
\bottomrule
\end{tabular}
\caption{Additional dialogue metrics.}
\label{tab:dynamic-supp}
\end{table*}

\section{Full Ablation Results}
\label{app:ablation}

Tables~\ref{tab:ablation-static} and~\ref{tab:ablation-dynamic} report the item-response and dialogue metrics of the full model and the three ablations. Removing the learner state gives the largest change in $W_1$ and $\mathrm{F1}_{\mathrm{struct}}$ on every dataset. Without the learner state, the standard deviation of simulated accuracy falls to 0.035--0.086, and on XES3G5M and ENEM it is larger with the persona prior than without it.

\begin{table*}[t]
\centering
\small
\setlength{\tabcolsep}{6pt}
\begin{tabular}{llccccc}
\toprule
& & & & & \multicolumn{2}{c}{Sim.\ acc.} \\
\cmidrule(lr){6-7}
Dataset & Variant & Agree$\uparrow$ & $W_1\downarrow$ & $\rho\uparrow$ & Mean & SD \\
\midrule
\multirow{4}{*}{XES3G5M}
& \methodname & 0.752 & 0.039 & 0.691 & 0.828 & 0.132 \\
& w/o Prior & 0.741 & 0.025 & 0.603 & 0.817 & 0.116 \\
& w/o State & 0.784 & 0.074 & 0.664 & 0.905 & 0.066 \\
& w/o Harness & 0.760 & 0.056 & 0.559 & 0.846 & 0.134 \\
\midrule
\multirow{4}{*}{Eedi}
& \methodname & 0.606 & 0.046 & 0.703 & 0.593 & 0.211 \\
& w/o Prior & 0.610 & 0.041 & 0.704 & 0.587 & 0.237 \\
& w/o State & 0.578 & 0.116 & 0.742 & 0.642 & 0.086 \\
& w/o Harness & 0.606 & 0.048 & 0.703 & 0.594 & 0.212 \\
\midrule
\multirow{4}{*}{ENEM}
& \methodname & 0.630 & 0.169 & 0.668 & 0.384 & 0.328 \\
& w/o Prior & 0.592 & 0.142 & 0.594 & 0.500 & 0.296 \\
& w/o State & 0.487 & 0.409 & 0.539 & 0.802 & 0.035 \\
& w/o Harness & 0.617 & 0.257 & 0.643 & 0.481 & 0.402 \\
\bottomrule
\end{tabular}
\caption{Item-response ablation results. Sim.\ acc.\ gives the mean and the standard deviation of simulated per-student accuracy. The real means are 0.831 (XES3G5M), 0.574 (Eedi), and 0.394 (ENEM).}
\label{tab:ablation-static}
\end{table*}

\begin{table*}[t]
\centering
\footnotesize
\setlength{\tabcolsep}{2.3pt}
\begin{tabular}{llcccccccc}
\toprule
Dataset & Variant & $\kappa_{\mathrm{func}}\uparrow$ & $\mathrm{MCC}_{\mathrm{func}}\uparrow$ & $\mathrm{BAcc}_{\mathrm{err}}\uparrow$ & $\mathrm{MCC}_{\mathrm{err}}\uparrow$ & $\mathrm{F1}_{\mathrm{err}}\uparrow$ & $\mathrm{Rec}_{\mathrm{struct}}\uparrow$ & $\mathrm{Prec}_{\mathrm{struct}}\uparrow$ & $\mathrm{F1}_{\mathrm{struct}}\uparrow$ \\
\midrule
\multirow{4}{*}{Eedi QATD2k}
& \methodname & 0.659 & 0.673 & 0.738 & 0.509 & 0.561 & 0.315 & 0.368 & 0.339 \\
& w/o Prior & 0.571 & 0.635 & 0.834 & 0.603 & 0.651 & 0.300 & 0.236 & 0.264 \\
& w/o State & 0.741 & 0.753 & 0.612 & 0.322 & 0.346 & 0.099 & 0.220 & 0.136 \\
& w/o Harness & 0.565 & 0.626 & 0.808 & 0.588 & 0.639 & 0.307 & 0.275 & 0.290 \\
\midrule
\multirow{4}{*}{Khan/CoMTA}
& \methodname & 0.370 & 0.390 & 0.660 & 0.267 & 0.464 & 0.565 & 0.249 & 0.346 \\
& w/o Prior & 0.326 & 0.359 & 0.618 & 0.200 & 0.426 & 0.578 & 0.224 & 0.323 \\
& w/o State & 0.382 & 0.436 & 0.519 & 0.063 & 0.148 & 0.077 & 0.262 & 0.119 \\
& w/o Harness & 0.348 & 0.367 & 0.625 & 0.212 & 0.443 & 0.563 & 0.249 & 0.346 \\
\bottomrule
\end{tabular}
\caption{Dialogue ablation results.}
\label{tab:ablation-dynamic}
\end{table*}

\section{Prompts}
\label{app:prompts}

Listings~\ref{lst:state} to~\ref{lst:check} give the system prompts that the dialogue pipeline uses for learner state inference, the predicted error, generation, and the harness check. In each request the prompt is followed by the visible context and the schema of the fields to return. The field names are the identifiers of the learner-state fields in Table~\ref{tab:fields}. In the prompts, the error locus follows Newman's error analysis \citep{newman1977analysis,white2010numeracy}, the error mechanism follows \citet{radatz1979error}, the faulty rule follows the procedural bugs of \citet{brown1978diagnostic}, and the engagement categories follow the ICAP framework \citep{chi2014icap}. The item-response pipelines use corresponding prompts for state inference, generation, and the check, with the fields of the item-response column of Table~\ref{tab:fields}.

\begin{lstlisting}[float=*tp,caption={Learner state inference.},label={lst:state}]
You are resolving the student's current local interaction state in third person.
You must not generate the student reply.
You must only infer the student's current state before the next reply is generated.
Do not use future turns, gold labels, official annotations, or evaluation targets.
If a persona prior summary is provided, use it only as a background prior for the current learner.
Do not treat the persona prior or retrieved exemplars as the answer script for the current turn.

Method-state definitions:
- competence_position: within_current_competence | within_zpd | beyond_zpd
- misconception_status: not_salient | active_misconception | destabilizing_misconception | resolved_misconception
- metacognitive_regulation_state: unmonitored | weak_monitoring | active_monitoring | self_correcting
- help_internalization_state: no_uptake | surface_uptake | partial_internalization | integrated_internalization
- help_seeking_state: help_avoidant | help_ambivalent | adaptive_help_seeking
- attempt_readiness: not_ready | ready_for_local_attempt | ready_to_revise
- engagement_form: exactly one of Passive | Active | Constructive | Interactive
- response_commitment_state: avoidant | tentative | committed
- response_maturity_cap: fragment_only | brief_local_explanation | limited_structured_reasoning | full_student_level_reasoning

Important interpretation rule:
- Do not treat mathematical incorrectness as a reason to default to tentative wording.
- A student can hold a misconception or give a wrong answer with high local commitment.
- Predict explicit uncertainty, clarification-seeking, or help-seeking only when it is supported by the visible dialogue state and tutor move.

- Every required string field must be non-empty.
- Every required evidence list must be non-empty.
- Each categorical field must be exactly one string value, never a list.
- Top-level keys must match required_schema exactly. Do not rename, omit, or merge fields.
Return only the requested JSON object.
\end{lstlisting}

\begin{lstlisting}[float=*tp,caption={Predicted error.},label={lst:error}]
You are building a third-person ErrorBlueprint for the student's next turn.
This is not free generation of the student reply.
Use only the visible dialogue history, the current tutor move, the current interaction state, and the predicted next state.
Do not use future turns, gold labels, official annotations, or evaluation targets.
If a persona prior summary is provided, use it only as a background prior for the current learner.
Do not treat the persona prior or retrieved exemplars as the answer script for the current turn.

Theory rules:
- For error locus, use Newman Error Analysis as summarized by White (2010):
  reading | comprehension | transformation | process_skills | encoding
- Choose the earliest evidenced breakdown in the Newman chain.
- For error mechanism, use Radatz (1979):
  language_difficulty | spatial_information_difficulty | deficient_prerequisites | incorrect_associations | irrelevant_rules
- For buggy_rule, use the procedural-bug idea from Brown & Burton:
  express one short executable mal-rule in plain language, not a broad personality description.

Local anchoring rules:
- Anchor the blueprint to the current tutor move, not to a random earlier error from the dialogue.
- wrong_claim_seed must target the exact local object the tutor is currently eliciting.
- Use tutor_turn_profile.elicited_response_target_type and tutor_turn_profile.elicited_object_description to decide what kind of wrong claim belongs here.
- If the tutor is asking for an option, judgment, or final local answer, the wrong_claim_seed should usually be an incorrect option, incorrect judgment, or incorrect local answer of that same kind.
- If the tutor is asking for the next step in a procedure, the wrong_claim_seed should usually be an incorrect local step at that exact stage, not a different earlier-stage mistake.
- Do not shift the student to a different subproblem or a different response function than the tutor's current question.

Newman interpretation rules for tutoring context:
- Use comprehension when the student is likely to misunderstand what the tutor is asking, what kind of answer is being requested, or how to interpret the compared methods/statements.
- Use transformation when the student is likely to choose the wrong representation, wrong operation, wrong equation, or wrong mathematical mapping for the current local task.
- Use process_skills when the tutor has already supplied the relevant representation, operation, or target form and the student is more likely to execute that local step incorrectly.
- Use encoding when the student has essentially completed the local work but is likely to express the final answer in the wrong written or symbolic form.

Error-presence rules:
- no_specific_error:
  use this when the next turn is not best represented as a specific mathematical error.
- non_codable_confusion:
  use this when the next turn is likely confused, incomplete, or help-seeking, but not stable enough to code as one specific mathematical error.
- codable_math_error:
  use this only when the next turn is likely to contain one concrete mathematical error that can be realized in the utterance.

Coding rules:
- For Brown & Burton style buggy rules, code the local executable wrong rule the student is most likely to apply next, not the corrected rule the tutor is trying to elicit.
- If the tutor has already supplied the correct representation or operation and is now checking execution, selection, or evaluation, do not drift back to a broader earlier-stage transformation error unless the next turn is truly likely to do that.
- If error_presence is not codable_math_error, set newman_error_locus, radatz_error_mechanism, buggy_rule, and wrong_claim_seed to null.
- If error_presence is codable_math_error, all four of those fields must be non-null.
- wrong_claim_seed must describe the local incorrect claim or incorrect step the student is most likely to say next, not a polished explanation.
- Keep the blueprint local to this turn. Do not script multiple future turns.
- Every required evidence list must be non-empty.
- Top-level keys must match required_schema exactly.
Return only the requested JSON object.
\end{lstlisting}

\begin{lstlisting}[float=*tp,caption={Generation of the turn.},label={lst:generation}]
You generate exactly one next student turn in a math tutoring dialogue.
You must stay within the predicted state and compiled constraints.
You must follow the single-step protocol and only finalize on the last allowed step.
This is the final allowed step, so step_kind must be exactly finalize_turn.
This final step must output the current student turn now, not another internal step.
Do not answer like a tutor.
Preserve the predicted uptake depth and misconception trajectory rather than jumping to sudden mastery.
Do not add uncertainty markers unless the predicted state or compiled constraints actually require them.
A student can be wrong while sounding locally direct.
If the compiled constraints require a concrete codable math error, the turn must contain a local incorrect claim consistent with the specified buggy rule and wrong-claim seed.
Do not jump ahead to future turns.
Do not use future turns, gold labels, official annotations, or evaluation targets.
Generate the way a student would actually say it in the moment, not a polished explanation of the student's state.
Keep the wording locally grounded, informal enough to sound like a student, and no more organized than the predicted maturity allows.
If the same content can be said in a more student-like way without changing meaning, choose the student-like wording.
Every required string field must be non-empty.
Top-level keys must match required_schema exactly. Do not rename, omit, or merge fields.
Return only the requested JSON object.
\end{lstlisting}

\begin{lstlisting}[float=*tp,caption={Harness check.},label={lst:check}]
You review whether a generated next student turn violates the predicted student state.
This is not answer grading.
This is state-consistency review only.

Focus on these theory-derived violations:
- over_fast_transition
- error_path_discontinuity
- over_mature_response

Pay special attention to:
- abrupt disappearance of the current misconception path
- replies that sound more internally resolved than the predicted attempt readiness allows
- cases where the reply invents new uncertainty or help-seeking that is not supported by the predicted local state
- cases where a predicted concrete codable error path disappears into generic confusion, generic hedging, or generic help-seeking

Every required string field must be non-empty.
Every required evidence list must be non-empty.
Top-level keys must match required_schema exactly. Do not rename, omit, or merge fields.
Return only the requested JSON object.
\end{lstlisting}

\end{document}